# BanglaByT5: Byte-Level Modelling for Bangla

**Pramit Bhattacharyya** and **Arnab Bhattacharya**
Dept. of Computer Science and Engineering,
Indian Institute of Technology Kanpur,
India
`{pramitb,arnabb}@cse.iitk.ac.in`

## Abstract

Large language models (LLMs) have achieved remarkable success across various natural language processing tasks. However, most LLM models use traditional tokenizers like BPE and SentencePiece, which fail to capture the finer nuances of a morphologically rich language like Bangla (Bengali). In this work, we introduce **BanglaByT5**, the first byte-level encoder-decoder model explicitly tailored for Bangla. Built upon a small variant of Google's ByT5 architecture, BanglaByT5 is pre-trained on a 14GB curated corpus combining high-quality literary and newspaper articles. Through zero-shot and supervised evaluations across generative and classification tasks, BanglaByT5 demonstrates competitive performance, surpassing several multilingual and larger models. Our findings highlight the efficacy of byte-level modelling for morphologically rich languages and highlight BanglaByT5 potential as a lightweight yet powerful tool for Bangla NLP, particularly in both resource-constrained and scalable environments.

## 1 Introduction

Large Language Models (LLMs) have redefined natural language processing (NLP) by achieving strong results across multiple tasks like machine translation, question answering, and paraphrasing. However, these models rely on subword tokenization (e.g., BPE, SentencePiece), which fragments words inconsistently and poses significant challenges when applied to *morphologically rich* Indian languages like Bangla (Brahma et al., 2025; Nehrdich et al., 2024). In contrast, *byte-level modelling* operates directly on raw bytes, enabling models to handle linguistic variations uniformly across scripts and domains.

In this paper, we introduce **BanglaByT5**, the first monolingual byte-level encoder-decoder model for Bangla, built on the small ByT5 architecture and pre-trained on a 14GB balanced corpus combining VĀCASPATI (literature) (Bhattacharyya et al., 2023) and IndicCorp (news) corpora (Kakwani et al., 2020). BanglaByT5 is evaluated on classification and generative tasks under zero-shot and fine-tuned settings. It outperforms similar-size models like IndicBART and BanglaT5, as well as BLOOM-1.1B (Scao et al., 2023), and performs within 5% of GPT2-XL (Radford et al., 2019) despite being 5 times smaller.

Our contributions are as follows:

- We propose **BanglaByT5**, the first monolingual byte-level encoder-decoder model for Bangla.
- We conduct extensive evaluation across tasks and settings, showing competitive or superior performance to larger models in resource-constrained scenarios.

## 2 Related Work

ELMo (Peters et al., 2018) and BERT (Devlin et al., 2019) have achieved strong results on NLU benchmarks. For generation tasks, decoder-only models like GPT-2/3 (Radford et al., 2019; Brown et al., 2020) and encoder-decoder models like T5 (Raffel et al., 2023), and mT5 (Xue et al., 2021)) have become prominent. ByT5 (Xue et al., 2022), a byte-level extension of mT5, has shown advantages for morphologically rich languages. While character-aware models like CharacterBERT (Boukkouri et al., 2020) and others (Ling et al., 2015; Chung et al., 2016; Jozefowicz et al., 2016; Wang et al., 2019; Wei et al., 2021; Kim et al., 2016) incorporate subword-free representations, they still rely on token boundaries. Other efforts (Garcia et al., 2021; Kudo, 2018) address tokenization challenges through vocabulary adaptation or randomized subword segmentation. Recent multilingual models like LLaMA-3 (Grattafiori et al., 2024), Mistral (Jiang et al., 2023), and IndicBART (Dabre et al., 2022) include Bangla. ByT5 model for morphologically rich language like Sanskrit (Nehrdich et al., 2024) has also been adopted. Monolingual models like BanglaT5 (Bhattacharjee et al., 2023) and Paramanu (Niyogi and Bhattacharya, 2024) are available for Bangla. Here, we propose BanglaByT5, the first byte-level monolingual Bangla model, based on a smaller variant of ByT5.

## 3 BanglaByT5

In this section we discuss in detail the various aspects of our proposed model, BanglaByT5.

### 3.1 Pretraining Data

We curated a corpus by merging the VĀCASPATI (Bhattacharyya et al., 2023) and IndicCorp (Kakwani et al., 2020) (Bangla corpus) for the pretraining the BanglaByT5 model. The merged corpus, 14GB in size, contains 947,041,525 words and 75,151,084 sentences with 12.60 words per sentence. Since IndicCorp is a newspaper dataset with ∼3.8 million articles and VĀCASPATI is entirely curated from literary data, we can assure the quality of the merged corpora, which is essential for training any GenAI model (Luccioni and Viviano, 2021). We have not used data from other sources such as websites or blogs since it was not feasible for us to ascertain the quality of such data. We have used the preprocessing steps mentioned in (Bhattacharyya et al., 2023) (Appx 8.1). The preprocessed corpus is used for the pretaining of BanglaByT5.

### 3.2 Pretraining Objective

ByT5, an encoder-decoder model, follows the same training objective as the original T5 model, specifically *span corruption denoising* task applied at byte level compared to token or subword level for T5. In ByT5, a fixed percentage of continuous byte spans are randomly selected and replaced with special sentinel tokens. The model is then trained to reconstruct the original spans, treating this as a sequence-to-sequence generation task. The byte-level denoising task enables the model to learn powerful representations without relying on language-specific tokenization. It allows for robust generalization, especially in morphologically rich languages like Bangla (Xue et al., 2022).

### 3.3 Model Architecture and Hyperparameters

Before the pretraining of the model on the merged corpus, we trained a byte-level tokenizer with 384 vocabulary size that includes 100 special tokens. This tokenizer generated 7,533,270,552 tokens for the merged corpus, resulting in a fertility score of 7.96, which is neither too high as Google-byt5-small (15.02) nor too small as BanglaT5 (1.20). The tokenizer is used for pretraining BanglaByT5.

We pre-trained the small variant of the Google-ByT5 model (Xue et al., 2022) with 12 hidden layers, 6 attention heads, 1472 hidden size, 3584 feed-forward size with gated-GELU activation (Shazeer, 2020).The model was trained with a batch size of 16 and a gradient accumulation step of 2 for over 3 million steps, utilizing two A100 40GB GPU instances. We employed the Adam optimizer (Kingma and Ba, 2017) with a learning rate of 3e-5, a linear warm-up for the first 500 steps, and a "cosine" learning rate scheduler. We train the model with a context size of 512 (∼5 sentences). The resulting model has ∼300M parameters yielding a token-to-parameter ratio of ∼25.14.

## 4 Evaluation

In this section, we explore the efficacy of BanglaByT5. We adopted a two-fold approach. First, we asked BanglaByT5 to generate responses to curated Bangla prompts to evaluate its generative abilities in the zero-shot setting. Zero-shot evaluation is particularly important because it reveals the model's inherent generative ability without reliance on domain-specific adaptation. Then, we evaluated the performance of BanglaByT5 on both classification and generation-based downstream tasks in supervised fine-tuning mode.

### 4.1 Prompt Generation and Multi-Turn Evaluation in Zero-Shot Setting

We adopted a two-stage evaluation methodology to assess the responses generated by BanglaByT5 and competing models. In the first stage, we evaluated the model's responses across four key dimensions using *LLaMa-3.1-8B* (Grattafiori et al., 2024) and *Mistral-7B* (Jiang et al., 2023) as LLM-as-a-Judge (Gu et al., 2025). The four key metrics used are *Fluency, Coherence, Relevance* and *Creativity* (definitions in Appx 8.3). Each prompt is run 5 times to capture the variation in generation by the LLMs and is graded on a scale of 1 to 10. Table 1 shows the performance of BanglaByT5 against other models. From the table, it is evident that the generation ability of BanglaByT5 is comparable to GPT2-XL (the best performing model) and is better than any other model even if it is twice (GPT2-Large) or thrice (BLOOM-1.1B (Scao et al.,

| Model | Params | Mistral-7B | | | | LLaMA-3.1-8B | | | |
|---|---|---|---|---|---|---|---|---|---|
| | | Fluency | Relevance | Coherence | Creativity | Fluency | Relevance | Coherence | Creativity |
| mt5-small | 240M | 1.60 ± 1.3 | 2.60 ± 1.6 | 2.50 ± 1.4 | 1.00 ± 1.4 | 2.00 ± 1.4 | 2.60 ± 1.5 | 2.50 ± 1.1 | 2.60 ± 1.8 |
| mt5-base | 580M | 6.00 ± 1.3 | 6.00 ± 1.2 | 6.60 ± 1.1 | 4.00 ± 1.6 | 6.00 ± 1.6 | 6.60 ± 1.5 | 6.00 ± 1.1 | 4.50 ± 1.6 |
| mt5-large | 1.1B | 8.60 ± 1.1 | 8.60 ± 1.3 | 8.60 ± 1.2 | 6.00 ± 1.4 | 8.60 ± 1.4 | 8.60 ± 1.6 | 8.60 ± 1.1 | 6.00 ± 1.6 |
| google-byt5-small | 300M | 6.60 ± 0.2 | 6.60 ± 0.4 | 6.30 ± 0.3 | 4.00 ± 0.6 | 6.60 ± 0.3 | 6.60 ± 0.2 | 6.00 ± 0.3 | 4.00 ± 0.7 |
| google-byt5-base | 580M | 7.60 ± 0.1 | 7.60 ± 0.6 | 7.60 ± 0.7 | 5.00 ± 0.6 | 7.60 ± 0.3 | 7.60 ± 0.5 | 7.00 ± 0.3 | 5.00 ± 0.7 |
| google-byt5-large | 1.2 B | 9.00 ± 0.2 | 9.00 ± 0.4 | 9.00 ± 0.3 | 6.00 ± 0.6 | 9.00 ± 0.3 | 9.00 ± 0.2 | 9.00 ± 0.3 | 6.00 ± 0.7 |
| GPT-2 Medium | 355M | 8.00 ± 0.1 | 7.00 ± 0.5 | 6.00 ± 0.3 | 6.00 ± 0.4 | 8.00 ± 0.2 | 7.00 ± 0.4 | 6.00 ± 0.2 | 6.00 ± 0.6 |
| GPT-2 Large | 774M | 9.00 ± 0.8 | 9.00 ± 0.4 | 8.00 ± 0.6 | 6.50 ± 0.5 | 9.00 ± 0.1 | 8.60 ± 0.8 | 8.00 ± 0.2 | 6.00 ± 0.4 |
| GPT-2 XL | 1.5B | **9.00 ± 0.2** | **9.00 ± 0.5** | **9.00 ± 0.5** | **6.50 ± 0.6** | **9.00 ± 0.2** | **9.00 ± 0.5** | **8.67 ± 0.2** | **7.00 ± 0.6** |
| BLOOM | 560M | 7.00 ± 0.3 | 6.50 ± 0.2 | 6.00 ± 0.4 | 4.00 ± 0.5 | 6.50 ± 0.2 | 6.50 ± 0.3 | 7.00 ± 0.4 | 4.00 ± 0.5 |
| BLOOM | 1.1B | 8.00 ± 0.3 | 7.70 ± 0.2 | 7.70 ± 0.3 | 5.00 ± 0.5 | 7.50 ± 0.2 | 8.00 ± 0.3 | 7.70 ± 0.4 | 5.50 ± 0.5 |
| IndicBART | 272M | 6.30 ± 0.4 | 7.00 ± 0.2 | 6.00 ± 0.3 | 3.00 ± 0.5 | 6.50 ± 0.1 | 7.30 ± 0.3 | 7.00 ± 0.4 | 3.50 ± 0.5 |
| BanglaT5 | 240M | 1.60 ± 1.1 | 3.00 ± 1.3 | 2.50 ± 1.2 | 1.00 ± 1.4 | 2.00 ± 1.4 | 3.60 ± 1.5 | 2.50 ± 1.1 | 2.80 ± 1.6 |
| Paramanu | 334M | 6.30 ± 0.4 | 7.00 ± 0.2 | 6.00 ± 0.3 | 3.00 ± 0.5 | 6.5 ± 0.1 | 7.30 ± 0.3 | 7.00 ± 0.4 | 3.50 ± 0.5 |
| BanglaByT5 | 300M | 8.60 ± 0.2 | 8.60 ± 0.4 | 8.30 ± 0.3 | 5.00 ± 0.6 | 8.60 ± 0.3 | 8.60 ± 0.2 | 8.00 ± 0.3 | 5.00 ± 0.7 |

Table 1: LLM evaluation of Bangla generation using Mistral-7B and LLaMA-3.1-8B as LLM-as-a-Judge.

| Model | Params | Sentiment | NER | MT (sacreBLEU) | Paraphrasing | GEC (GLEU) |
|---|---|---|---|---|---|---|
| mt5-small | 240M | 62.50 ± 1.35 | 28.10 ± 1.42 | 20.10 ± 1.43 | 27.80 ± 1.56 | 68.00 ± 1.47 |
| mT5-Base | 580M | 67.50 ± 1.35 | 33.10 ± 1.45 | 23.10 ± 1.43 | 30.50 ± 1.57 | 70.00 ± 1.87 |
| ByT5-small | 300M | 64.60 ± 0.20 | 30.60 ± 0.55 | 21.86 ± 0.50 | 29.28 ± 1.60 | 68.00 ± 1.45 |
| ByT5-base | 580M | 67.60 ± 0.20 | 32.80 ± 0.55 | 23.86 ± 0.50 | 30.48 ± 1.60 | 71.10 ± 1.45 |
| GPT-2 Medium | 355M | 63.00 ± 1.50 | 29.00 ± 1.20 | 22.00 ± 1.50 | 29.00 ± 1.70 | 68.00 ± 1.40 |
| GPT-2 Large | 774M | 67.80 ± 1.44 | **35.00 ± 1.35** | 24.20 ± 1.62 | **31.50 ± 1.60** | 71.20 ± 1.50 |
| BLOOM | 560M | 64.90 ± 1.47 | 31.50 ± 1.39 | 22.80 ± 1.63 | 29.60 ± 1.62 | 68.20 ± 1.55 |
| IndicBART | 272M | 63.40 ± 1.45 | 30.80 ± 1.36 | 22.40 ± 1.60 | 29.40 ± 1.56 | 67.50 ± 1.50 |
| BanglaT5 | 240M | 67.80 ± 1.40 | 33.00 ± 1.30 | 22.50 ± 1.50 | 29.80 ± 1.60 | 69.50 ± 1.50 |
| Paramanu | 334M | 66.00 ± 1.40 | 32.20 ± 1.30 | 21.90 ± 1.40 | 28.50 ± 1.50 | 68.70 ± 1.50 |
| **BanglaByT5** | 300M | **68.30 ± 0.20** | 33.60 ± 0.35 | **24.36 ± 1.50** | 31.28 ± 1.60 | **71.27 ± 1.40** |

Table 2: Scores of each Bangla model on five tasks.

2023)) in size.

We further evaluated the generation ability of BanglaByT5 model by comparing its responses for the 2000 prompts against the responses by two widely used reference models **LLaMA-3.1-8B** and **Mistral-7B**. We further benchmarked the response generated by BanglaByT5 by comparing it with the competing models. Each model was executed 5 times, and the mean and standard deviations are shown in Table 3. Since the responses for the reference models varied across the five runs, this approach enabled us to accurately assess the performance of BanglaByT5 in comparison to LLaMa-3.1-8B and Mistral-7B.

Table 1 and Table 3 indicate that the generation ability of BanglaByT5 is better than models with twice to thrice its parameter size and is comparable to GPT2-XL, which is five times larger. This generation ability of BanglaByT5 makes it a suitable model for deployment in low-resource settings. In detail, we have discussed the deployment potential of BanglaByT5 in Appx 8.5.

### 4.2 Supervised Fine Tuning

We further investigated the performance BanglaByT5 on various downstream tasks and benchmarked it against similar and larger parameter models on both classification and generation tasks.

**Sentiment Classification:** We used the dataset curated by (Islam et al., 2018), which comprises 3 polarity labels, positive, negative, and neutral, and is collected from social media comments on news and videos covering 13 domains, including politics, education, and agriculture. It consists of 5,709 negative, 6,410 positive, and 3,609 neutral sentences. For this classification task, we used *macro-F1* as an evaluation metric.

**NER:** We chose the publicly available Naamapadam (Mhaske et al., 2023) (Bengali subset) for this classification task. The dataset consists of 961.7K sentences for training, and 4.9K sentences have been used for evaluation. The tokens are tagged into 4 classes: Person (Per), Location (Loc), Organization (Org), and other (O). We have used *macro-F1* as an evaluation metric for the task.

**Machine Translation:** Machine Translation (MT) is one of the most studied generative tasks in

| Model | Params | LLaMA-3.1-8B | | | Mistral-7B | | |
|---|---|---|---|---|---|---|---|
| | | BERTScore | BLEU | METEOR | BERTScore | BLEU | METEOR |
| MT5-SMALL | 300M | 49.31 ± 1.40 | 1.50 ± 1.20 | 1.40 ± 1.30 | 48.89 ± 1.56 | 1.20 ± 1.10 | 1.32 ± 1.20 |
| MT5-BASE | 580M | 49.56 ± 1.60 | 1.56 ± 1.40 | 1.45 ± 1.30 | 49.99 ± 1.78 | 1.65 ± 1.42 | 1.74 ± 1.30 |
| MT5-LARGE | 1.2B | 67.03 ± 1.70 | 19.29 ± 1.90 | 37.37 ± 1.60 | 61.66 ± 1.58 | 9.90 ± 1.86 | 23.22 ± 1.54 |
| BYT5-SMALL | 300M | 71.10 ± 1.50 | 1.71 ± 1.30 | 13.17 ± 1.65 | 70.32 ± 1.42 | 1.19 ± 1.23 | 12.71 ± 1.56 |
| BYT5-BASE | 580M | 71.32 ± 1.50 | 3.54 ± 1.60 | 14.27 ± 1.55 | 70.56 ± 1.45 | 2.57 ± 1.35 | 16.15 ± 1.58 |
| BYT5-LARGE | 1.2B | 75.80 ± 1.60 | 8.66 ± 1.80 | 21.20 ± 1.63 | 74.23 ± 1.55 | 7.54 ± 1.70 | 19.15 ± 1.56 |
| BLOOM-560M | 560M | 72.49 ± 1.40 | 6.96 ± 1.50 | 32.84 ± 1.68 | 72.43 ± 1.45 | 8.52 ± 1.56 | 30.61 ± 1.52 |
| BLOOM-1B | 1.1B | 72.63 ± 1.50 | 7.23 ± 1.60 | 33.45 ± 1.52 | 74.68 ± 1.65 | 9.46 ± 1.64 | 31.50 ± 1.54 |
| GPT-2 MEDIUM | 355M | 76.08 ± 1.60 | 11.76 ± 1.50 | 33.56 ± 1.64 | 75.00 ± 1.54 | 8.26 ± 1.52 | 30.45 ± 1.44 |
| GPT2-LARGE | 774M | 80.51 ± 1.60 | 31.26 ± 1.70 | 50.92 ± 1.50 | 80.40 ± 1.64 | 30.66 ± 1.61 | 49.25 ± 1.54 |
| GPT2-XL | 1.5B | **81.69 ± 1.60** | **32.40 ± 1.80** | **51.56 ± 1.50** | **81.29 ± 1.56** | **31.86 ± 1.60** | **50.46 ± 1.56** |
| IndicBART | 272M | 61.81 ± 1.60 | 1.86 ± 1.40 | 5.40 ± 1.57 | 62.58 ± 1.44 | 1.86 ± 1.35 | 4.75 ± 1.46 |
| BanglaT5 | 240M | 63.81 ± 1.60 | 2.86 ± 1.45 | 7.49 ± 1.57 | 64.08 ± 1.42 | 2.86 ± 1.30 | 6.75 ± 1.40 |
| Paramanu | 334M | 62.31 ± 1.40 | 2.00 ± 1.72 | 6.00 ± 1.56 | 62.88 ± 1.45 | 2.26 ± 1.38 | 6.25 ± 1.45 |
| BanglaByT5 | 300M | 78.21 ± 1.10 | 12.41 ± 1.96 | 34.08 ± 1.61 | 75.84 ± 1.03 | 9.06 ± 1.38 | 31.63 ± 1.54 |

Table 3: Benchmarking the generation ability of BanglaByT5 models in zero-shot setting.

Bangla. For this task, we curated a dataset by merging the dataset created by Gala et al. (2023) (1,022 sentences) and Costa-jussà et al. (2022) (3,001 sentences) and creating a comprehensive dataset of 4,023 sentences. We used 80% of the data for training the model and tested on the remaining 20%. We used the *sacreBLEU* score as the evaluation metric for the task.

**Paraphrasing:** Paraphrasing refers to the rephrasing of a sentence or passage using different words and structures while preserving its original meaning. We used the publicly available paraphrasing dataset curated by Akil et al. (2022) for this task. The dataset consists of 5,763 sentences, of which 80% is used for training and the remaining 20% for testing. We used the *sacreBLEU* score as the evaluation metric for the task.

**Grammatical Error Correction:** Grammatical Error Detection (GEC) refers to the task of automatic detection and correction of grammatical errors in a sentence. It is one of the most important generative tasks as it also tests the model's understanding of generating semantically correct sentences. We used the VAIYAKARANA dataset curated by Bhattacharyya and Bhattacharya (2024). The dataset consists of 111,256 sentences divided into 12 finer classes. Like the other generative tasks, we have used 80% of the dataset for training and 20% for testing. We used *GLEU* as the evaluation metric for the task.

### 4.2.1 Results

We compared the performance of BanglaByT5 on the specified downstream tasks against similar parameter models like mT5-small, mT5-base, google-ByT5-small, google-ByT5-base, BanglaT5, IndicBart, Paramanu, gpt2-medium and gpt2-large. All the pre-trained models are run between 5-25 epochs on a single instance of Nvidia A100-46 GB GPU. We have used beam search for inferencing (using 10 beams) and set the temperature at 0.7 and the top_k value at 70. The maximum output length has been set at 512 for all the models. Table 2 shows the result of all the models on the downstream tasks based on the evaluation metrics discussed in this section. Table 2 shows that BanglaByT5 outperforms all similar parameter models on generative tasks such as MT, paraphrasing and GEC while giving comparable results on classification tasks like Sentiment classification and NER. BanglaByT5 also performs similarly to the GPT2-Large model on all the generative tasks, outperforming it on classification tasks. The results indicate the efficacy of BanglaByT5 as a generative model for Bangla. Additionally, we have benchmarked the performance of BanglaByT5 against larger models like google-byt5-large, mT5-large, GPT2-XL and BLOOM-1.1B, results of which are shown in Table A2 of Appx 8.4. BanglaByT5 outperforms BLOOM-1.1B in all tasks and performs within 5% of the other larger models.

## 5 Conclusions

We presented **BanglaByT5**, a byte-level monolingual language explicitly tailored for morphologically rich languages like Bangla. Through rigorous evaluation, we demonstrate that BanglaByT5 surpasses existing Bangla models and matches or outperforms larger multilingual systems in both generation and classification tasks. Moreover, its scalability in low-resource environments positions it as a practical and impactful tool for Bangla NLP. This work underscores the potential of byte-level modelling in building robust LLMs for morphologically rich languages.

## 6 Limitations

**Lack of large quantity of quality data:** Bangla inherently suffers from large quantity of quality data. We have been able to curate only 14GB of data, prompting us to use a small variant of google-ByT5. Our results indicate that a larger variant pre-trained over a large quality corpus will benefit Bangla.

**Hallucination:** Hallucination is an inherent property of any LLMs (Xu et al., 2025). Hence, we cannot always guarantee the factual correctness of responses generated by BanglaByT5.

**Memorization:** Like hallucination *memorization* is also an inherent ability of LLMs (Hartmann et al., 2023). However, Carlini et al. (2019) showed that models with <=300M parameters show minimal memorization. Hence, we have not tested the memorization ability of BanglaByT5.

## 7 Ethics Statement

The corpus used for pre-training BanglaByT5 is curated by merging IndicCorp (Kakwani et al., 2020) and VĀCASPATI (Bhattacharyya et al., 2023). The authors of VĀCASPATI have provided us with the corpus, and IndicCorp is publicly available. Hence, there is no copyright infringement in the curation of the merged corpus. Since IndicCorp is a newspaper corpus and VĀCASPATI is a literary corpus, there will be minimal chances of having objectionable and offensive statements. For Grammar Error Correction (GEC) work, the authors of VAIYAKARANA also provided us with the dataset. Hence, there is no copyright infringement. We will release BanglaByT5 and the pre-trained dataset upon acceptance of the paper under a non-commercial license.

**Carbon Footprint:** We estimate the carbon emissions incurred during the pretraining of BanglaByT5 on 2 NVIDIA A100 (40GB) GPUs, each with a Thermal Design Power (TDP) of 250W, for 600 training hours. This results in an energy consumption of approximately $0.25 kW \times 2 \times 600 hours = 300 kWh$. Assuming an average carbon intensity of $0.7\ kgCO_2/kWh$, the total carbon emission is estimated as:

$$300 kWh \times 0.7\ \frac{kgCO_2 kWh}{=}\ \mathbf{210\ kgCO_2}$$

which is significantly lower than the emissions reported for large-scale models such as GPT2-XL (Strubell et al., 2019).

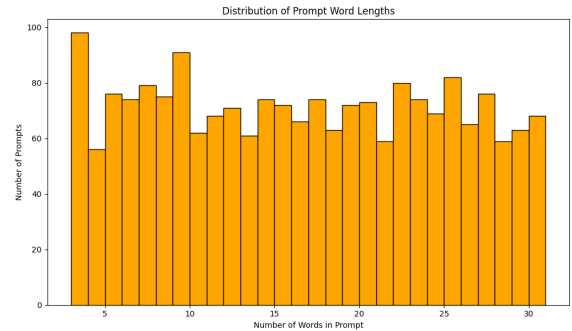

Figure 1: Figure showing distribution of prompt length by word.

## 8 Appendix

### 8.1 Data Cleaning

- *Cleaning of Unicode characters*: Unicode characters "$0020$" (space), "$00a0$" (no-break space), "$200c$" (zero width non-joiner), "$1680$" (ogham space mark), "$180e$" (mongolian vowel separator), "$202f$" (narrow no-break space), "$205f$" (medium mathematical space), "$3000$" (ideographic space), "$2000$" (en quad), "$200a$" (hair space) are removed from the texts.
- *Cleaning of different punctuation marks*: Usage of punctuation marks have also evolved alongside words in Bangla. In total we have removed all 36 types of Bangla punctuation marks.

### 8.2 Prompt generation

Table A1 shows the number of distribution of words in curated prompts.

| Word Length | Prompt Count |
|---|---|
| 3 | 98 |
| 4 | 56 |
| 5 | 76 |
| 6 | 74 |
| 7 | 79 |
| 8 | 75 |
| 9 | 91 |
| 10 | 62 |
| 11 | 68 |
| 12 | 71 |
| 13 | 61 |
| 14 | 74 |
| 15 | 72 |
| 16 | 66 |
| 17 | 74 |
| 18 | 63 |
| 19 | 72 |
| 20 | 73 |
| 21 | 59 |
| 22 | 80 |
| 23 | 74 |
| 24 | 69 |
| 25 | 82 |
| 26 | 65 |
| 27 | 76 |
| 28 | 59 |
| 29 | 63 |
| 30 | 68 |

Table A1: Number of prompts by word length (3 to 30 words).

### 8.3 Evaluation Metrics for Zero-shot Evaluation

BanglaByT5 generation ability have been evaluated using four metrics keeping LLaMa-3.1 (8B) and Mistral-7B.

- **Fluency** — It refers to the grammatical correctness and naturalness of the generated language.
- **Coherence** — Coherence signifies the consistency and structure of multi-turn responses.
- **Relevance** — Relevance refers to the contextual alignment with the original prompt.
- **Creativity** — Creativity is defined as the novelty and expressiveness of the generated response.

### 8.4 Benchmarking against larger models

In this section we benchmarked BanglaByT5 against larger models like google-ByT5-large, mT5-large, GPT2-XL and BLOOM-1.1B on the downstream tasks specified in Sec 4.2. BanglaByT5 outperforms BLOOM-1.1B on all tasks and perform with 2-5% of the other models in spite of being 4-5 times smaller. Table A2 shows the result of BanglaByT5 and the larger models.

### 8.5 Deployability

In this section, we evaluate the scalability of the ByT5 model under CPU-only and GPU-accelerated environments and compare them with similar parameter models like google-byT5, google-mT5, banglaT5 and others to assess the deployment potential of the ByT5 model. *GPU-scalability* reflects how well the model leverages parallelism for high-throughput or real-time deployment, while *CPU-scalability* captures performance in low-resource environments. This dual perspective is essential for understanding the potential of deploying the ByT5 model in cloud and offline settings.

**Latency** denotes the average time (in seconds) required to generate an output of a single prompt, including tokenization, model forward pass and decoding. **Throughput**, on the other hand, focussed on number of prompts processed per second. We also monitor the peak memory usage for a prompt in both CPU and GPU mode, as this is a critical consideration for deployment. To evaluate the deployment potential, we curated a dataset of 200 prompts for the paragraph generation task with varying word lengths (5-25), with an average of 9.81 words per prompt, similar to the average word length found in Bangla (()). The detailed distribution of the number of prompts by varying word length is shown in Fig 2.

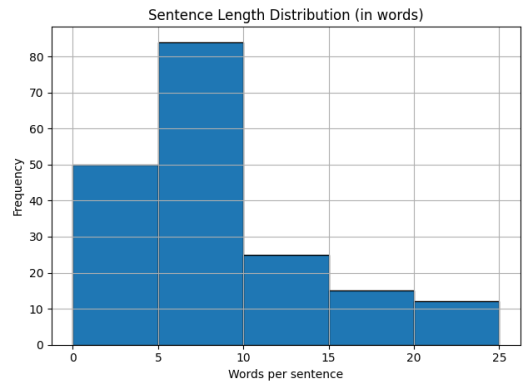

Figure 2: Figure showing distribution of word size in 200 prompts used for latency and throughput testing

Table A3 shows the variation in latency, throughput and memory required in cpu-only mode with an increase in batch size. From the table, it is seen that latency decreases and throughput increases with an increase in batch size, which is the ideal scenario. The peak memory usage is ∼2744MB (2.68GB). Hence, the model can be deployed in an offline system with 4 GB of RAM.

| Model | Params | Sentiment | NER | MT (sacreBLEU) | Paraphrasing | GEC (GLEU) |
|---|---|---|---|---|---|---|
| mT5-Large | 1.2B | 69.60 ± 1.30 | 35.00 ± 1.40 | 25.30 ± 1.65 | 32.73 ± 1.50 | 74.32 ± 1.40 |
| Google-ByT5-Large | 1.2B | 70.90 ± 1.40 | 36.90 ± 1.50 | 26.35 ± 1.62 | 33.54 ± 1.50 | 74.88 ± 1.40 |
| GPT2-XL | 1.5B | **72.47 ± 1.40** | **37.30 ± 1.56** | **28.58 ± 1.72** | **33.84 ± 1.56** | **75.83 ± 1.50** |
| BLOOM-1.1B | 1.1B | 66.0 ± 1.45 | 31.50 ± 1.35 | 23.50 ± 1.55 | 30.60 ± 1.45 | 70.20 ± 1.53 |
| **BanglaByT5** | 300M | 68.30 ± 0.20 | 33.60 ± 0.35 | 24.36 ± 1.5 | 31.28 ± 1.6 | 71.27 ± 1.4 |

Table A2: Performance comparison of BanglaByT5 against larger models on five Bangla NLP tasks.

| Batch | Latency | Throughput | Memory(MB) |
|---|---|---|---|
| 1 | 0.5646 | 1.77 | 2216.37 |
| 2 | 0.2692 | 3.71 | 2330.37 |
| 4 | 0.155 | 6.45 | 2418.31 |
| 8 | 0.0949 | 10.54 | 2582.26 |
| 16 | 0.0855 | 11.7 | 2601.56 |
| 32 | 0.0828 | 12.07 | 2742.99 |
| 64 | 0.0806 | 12.41 | 2743.99 |

Table A3: CPU-only scalability results for ByT5 across increasing batch sizes.

| Batch | Latency (s) | Throughput | GPU-Mem (MB) |
|---|---|---|---|
| 1 | 1.0927 | 0.92 | 1166.09 |
| 2 | 0.1592 | 6.28 | 1177.29 |
| 4 | 0.1296 | 7.72 | 1192.34 |
| 8 | 0.0810 | 12.34 | 1238.53 |
| 16 | 0.0439 | 22.80 | 1322.05 |
| 32 | 0.0409 | 24.48 | 1487.59 |
| 64 | 0.0146 | 68.26 | 1812.68 |

Table A4: GPU scalability results for ByT5 across batch sizes.

Tab A4 shows the variation in latency and throughput along with CPU and GPU requirements in gpu-available mode on the same 200 prompts. The maximum GPU requirement is ∼1.77GB. Further analysis shows that maximum cpu-requirement is ∼588 MB when batch size is 1.

Table A3 and Table A4 demonstrate that GPU acceleration yields substantial gains in throughput and reduces latency per sentence, but GPU memory usage increases sharply with batch size. In contrast, CPU-based inference falls behind in throughput but remains viable for offline deployments, especially in systems with limited memory.